\documentclass[a4paper, conference]{IEEEtran}
\usepackage{cite}
\usepackage{amsmath,amssymb,amsfonts}
\usepackage{algorithmic}
\usepackage{graphicx}
\usepackage{array}
\usepackage{arydshln}
\usepackage{textcomp}
\usepackage{makecell}
\usepackage{multirow}
\usepackage{xcolor}
\usepackage{graphicx}
\usepackage{subfigure}
\usepackage{caption}

\usepackage[font={footnotesize}]{caption}

\usepackage{balance}

\usepackage[utf8]{inputenc}

\usepackage[T1]{fontenc}

\def\BibTeX{{\rm B\kern-.05em{\sc i\kern-.025em b}\kern-.08em
    T\kern-.1667em\lower.7ex\hbox{E}\kern-.125emX}}
\renewcommand{\figurename}{Figure}

\makeatletter
\def\ps@IEEEtitlepagestyle{%
  \def\@oddfoot{\mycopyrightnotice}%
  \def\@evenfoot{}%
}
\def\mycopyrightnotice{%
  \footnotesize
  \vspace*{40pt}
  \parbox{\textwidth}{%
    \centering
    \fbox{%
      \begin{minipage}{0.9\textwidth}
        © 2024 IEEE.  Personal use of this material is permitted.  Permission from IEEE must be obtained for all other uses, in any current or future media, including reprinting/republishing this material for advertising or promotional purposes, creating new collective works, for resale or redistribution to servers or lists, or reuse of any copyrighted component of this work in other works.
      \end{minipage}%
    }%
  }%
  \gdef\mycopyrightnotice{}
}
\newcolumntype{C}[1]{>{\centering\let\newline\\\arraybackslash\hspace{0pt}}m{#1}}

\DeclareRobustCommand*{\IEEEauthorrefmark}[1]{%
\raisebox{0pt}[0pt][0pt]{\textsuperscript{\footnotesize\ensuremath{#1}}}}

\begin{document}

\title{Addressing Class Imbalance and Data Limitations in Advanced Node Semiconductor Defect Inspection: A Generative Approach for SEM Images\\
}

\author{\IEEEauthorblockN{Bappaditya Dey\IEEEauthorrefmark{1}*,
Vic De Ridder\IEEEauthorrefmark{1}$^,$\IEEEauthorrefmark{2}*,
 Victor Blanco\IEEEauthorrefmark{1},
Sandip Halder\IEEEauthorrefmark{1} \& Bartel Van Waeyenberge\IEEEauthorrefmark{2}}
\IEEEauthorblockA{\IEEEauthorrefmark{1}
Interuniversity Microelectronics Centre (imec), 3001 Leuven, Belgium}
\IEEEauthorblockA{\IEEEauthorrefmark{2}
Ghent University,9000 Ghent, Belgium}

*Vic De Ridder and Bappaditya Dey contributed equally to the work. \\
{\it Bappaditya.Dey@imec.be}}

\maketitle

\begin{abstract}
As semiconductor technology advances in accordance with Moore's Law, the continuous reduction in feature sizes presents an escalating challenge in defect detection. The limitations imposed by optical and e-beam technologies further contribute to the complexity of identifying defects at these shrinking scales, owing to increasing noise and decreasing image contrast levels. Precision in identifying nanometer-scale device-killer defects is crucial in both semiconductor research and development as well as in production processes. The effectiveness of existing machine learning-based approaches in this context is largely limited by the scarcity of data, as the production of real semiconductor wafer data for training these models involves high financial and time costs. Moreover, the existing simulation methods fall short of replicating images with identical noise characteristics, surface roughness and stochastic variations at advanced semiconductor nodes. In this paper, we propose a method for generating synthetic semiconductor SEM images using a diffusion model within a limited data regime. In contrast to images generated through conventional software simulation methods, SEM images generated through our proposed deep learning method closely resemble real SEM images, replicating their noise characteristics and surface roughness adaptively. Our main contributions, which are validated on three different real semiconductor datasets, are: i) proposing a patch-based generative framework utilizing DDPM to create SEM images with intended defect classes, addressing challenges related to class-imbalance and data insufficiency, ii) demonstrating generated synthetic images closely resemble real SEM images acquired from the tool, preserving all imaging conditions and metrology characteristics without any metadata supervision, iii) demonstrating a defect detector trained on generated defect dataset, either independently or combined with a limited real dataset, can achieve similar or improved performance on real wafer SEM images during validation/testing compared to exclusive training on a real defect dataset, iv) demonstrating the ability of the proposed approach to transfer defect types, critical dimensions, and imaging conditions from one specified CD/Pitch and metrology specifications to another, thereby highlighting its versatility.
\end{abstract}


\section{Introduction}
\label{sec:intro}

\begin{figure*}[t!]
    \centering
    \begin{minipage}{\textwidth}
    \begin{minipage}{0.370\linewidth}
        \includegraphics[width=\linewidth]{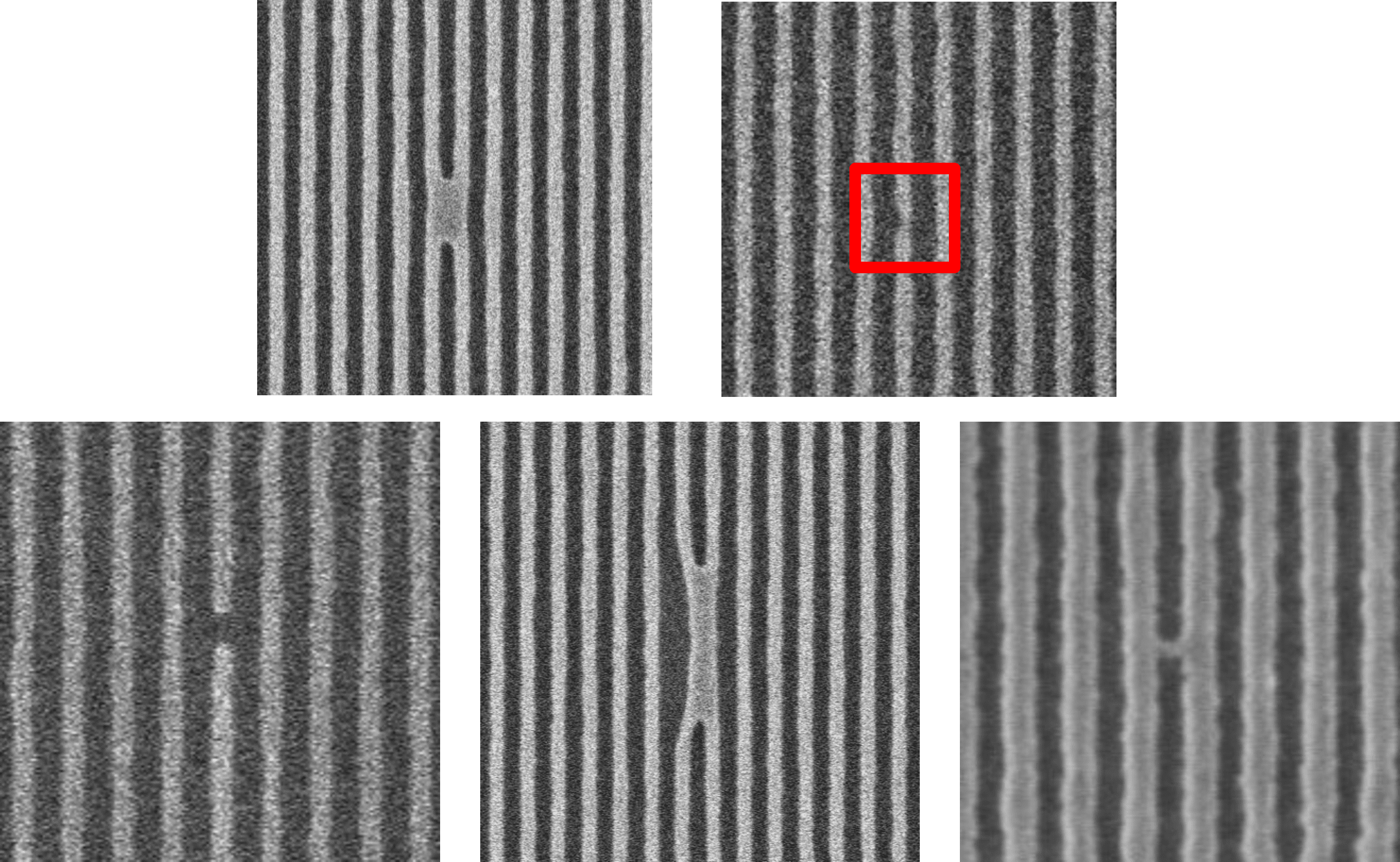}
        \caption*{(a) LS-ADI}
        \end{minipage}
        \hfill
        \begin{minipage}{0.245\linewidth}
            \includegraphics[width=\linewidth]{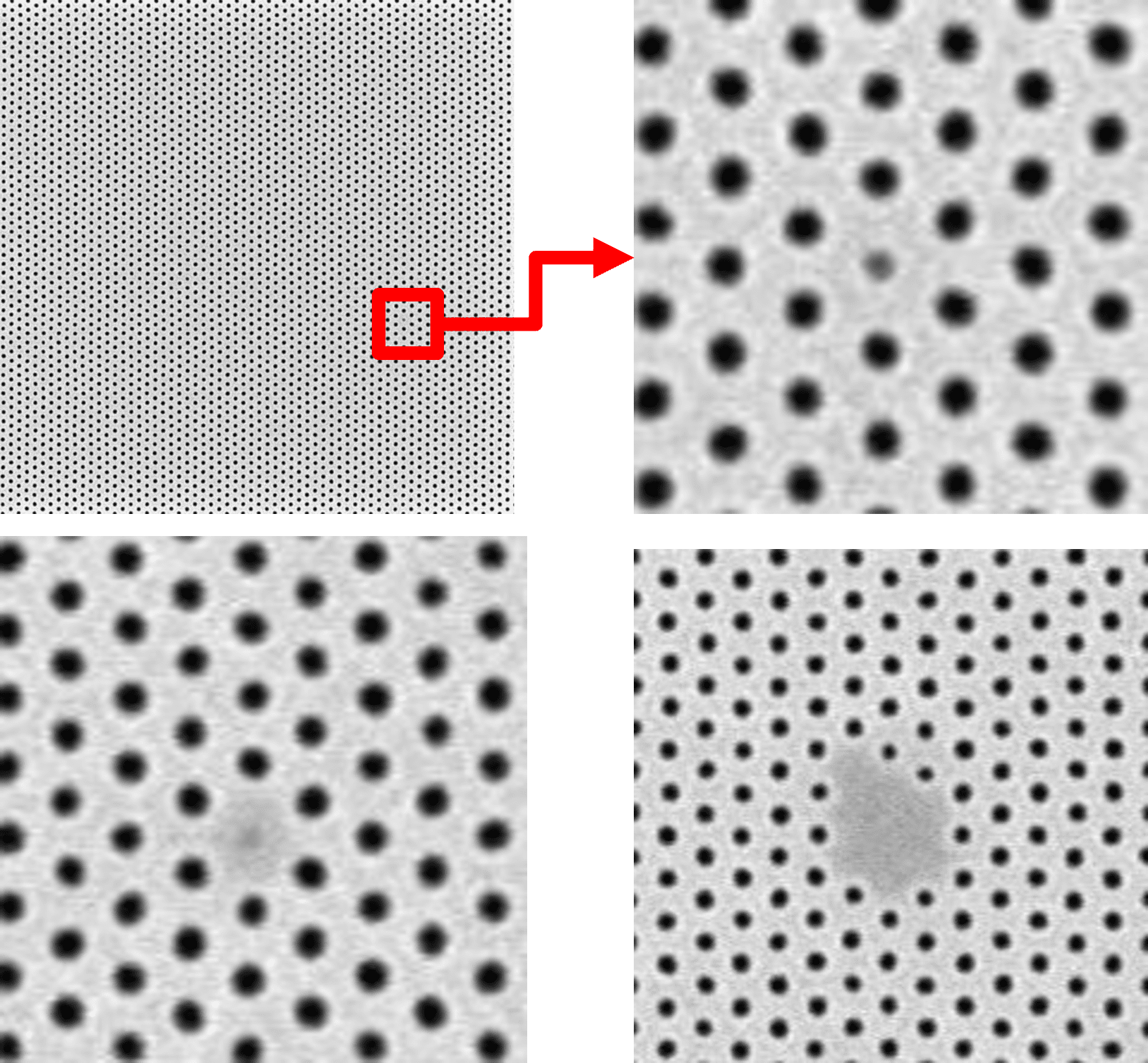}
                        \caption*{(b) HEXCH-DSA}
        \end{minipage}
        \hfill
        \begin{minipage}{0.360\linewidth}
            \includegraphics[width=\linewidth]{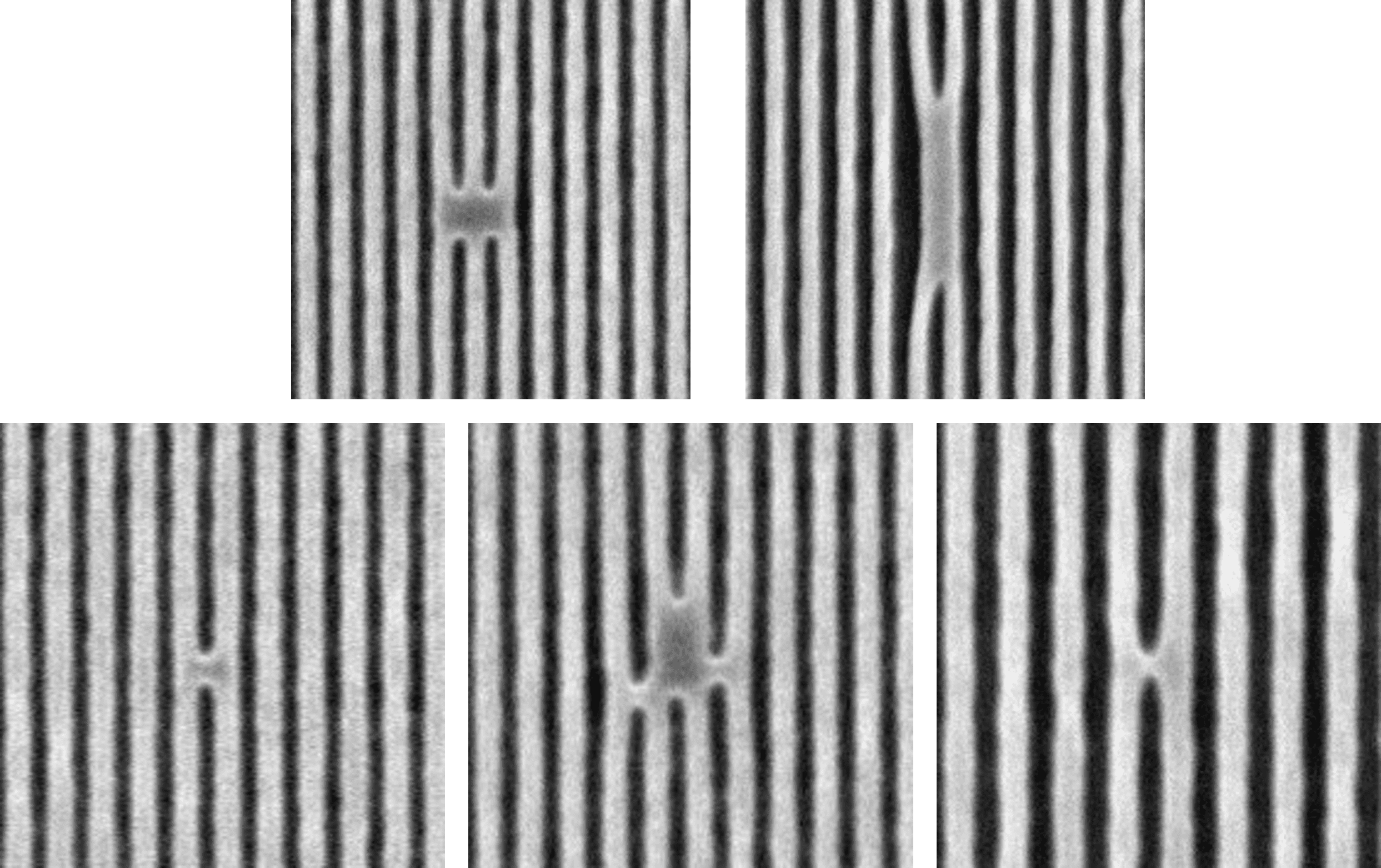}
            \caption*{(c) LS-AEI}
        \end{minipage}
        \caption{Representative sample SEM images illustrating example defect types in the datasets used in this study }
        \label{exampledataset}
    \end{minipage}
\end{figure*}

As Moore's Law drives the semiconductor industry towards achieving ever-smaller feature sizes ($\leq$10nm) and increased transistor density, the traditional methods of patterning face challenges. New approaches, including emerging lithography technologies like Extreme-Ultra-Violet-Lithography (EUVL) ($\leq$7nm), high-NA EUVL ($\leq$2nm), and other alternatives, are (being) developed to keep pace with Moore's Law and maintain the relentless pursuit of smaller feature sizes. The escalating complexity of semiconductor devices necessitates a corresponding elevation in process control. This entails the integration of precise metrology, sophisticated data analysis, and cutting-edge defect inspection methodologies. The prevailing state-of-the-art (SOTA) defect detection tools, whether optical or e-beam based, exhibit specific limitations. These tools rely on rule-based techniques for defect classification and detection, which introduces constraints in their adaptability and effectiveness. The use of rule-based approaches implies that these tools are programmed with predefined criteria to identify and classify defects. While this methodology is effective for well-understood and predictable defect patterns, it becomes increasingly challenging when dealing with complex, evolving, or stochastic defects \cite{de2017stochastic}, specifically in the presence of reduced signal-to-noise ratio (SNR) and image contrast.

Due to the inadequacy of rule-based methods at advanced nodes \cite{rulesfail}, DL-based object detectors have emerged as the state-of-the-art for stochastic defect inspection \cite{kondo2021massive}. However, the acquisition of a relevant stochastic defect dataset for training ML models faces considerable challenges within the semiconductor manufacturing domain. Not only is such a dataset rare and inherently noisy, but its acquisition is also a costly endeavor. The rarity of stochastic defect instances makes it challenging to compile a comprehensive dataset that accurately represents the diverse range of stochastic defects encountered in real-world semiconductor manufacturing processes.
Additionally, two significant bottlenecks further complicate the use of stochastic defect datasets in semiconductor manufacturing defect detection: (a) class imbalance, which arises when certain defect types are underrepresented or occur infrequently in the dataset, leading to a skewed distribution. This imbalance can compromise the model's ability to generalize and accurately detect defects across all classes. (b) insufficient dataset size,  as a limited amount of data may not adequately capture the variability and complexity of stochastic defects. The inherent diversity of semiconductor manufacturing processes demands large and representative datasets to ensure the robust training of machine learning models.
Addressing these challenges requires innovative approaches to dataset acquisition, including strategic data augmentation techniques to enhance dataset diversity. Collaboration within the industry and the development of shared datasets could also contribute to mitigating the issues associated with rare, noisy, and expensive stochastic defect datasets. Overcoming these challenges is pivotal for advancing the capabilities of machine learning models in semiconductor manufacturing defect detection.

In this research work, we use Denoising Diffusion Probabilistic Models (DDPM) to generate realistic semiconductor wafer SEM images, thereby increasing defect inspection training data and improving defect inspection performance. Our main contributions are: 

i) we propose a patch-based generative framework utilizing DDPM to generate SEM images that include intended defect classes with randomly variable instances, aiming to address class-imbalance and data insufficiency bottlenecks. This approach leads to an enhancement in defect detection performance, particularly in terms of precision and recall.

ii) our proposed approach generates synthetic images that closely resemble real ones, preserving actual characteristics without the need for prior knowledge of imaging settings (Best-Known-Methods).

 iii) we demonstrated that a defect detector trained on a generated defect dataset, either independently or in combination with a limited real dataset, can achieve a similar or improved mAP on real wafer SEM images during validation/testing compared to when trained exclusively on a real defect dataset. This trend was consistent across three different SEM datasets, validating the capability of DDPM to generate images with characteristics identical to real SEM images.

Finally, iv) our proposed approach demonstrates the capability to transfer defect types, critical dimensions, and imaging conditions from one specified CD/Pitch and metrology specifications to another CD/Pitch and metrology specifications.
\section{Related Works: Semiconductor Defect Inspection}
\label{sec:formatting}

\begin{figure*}[t!]
    \centering
    \includegraphics[width=\linewidth]{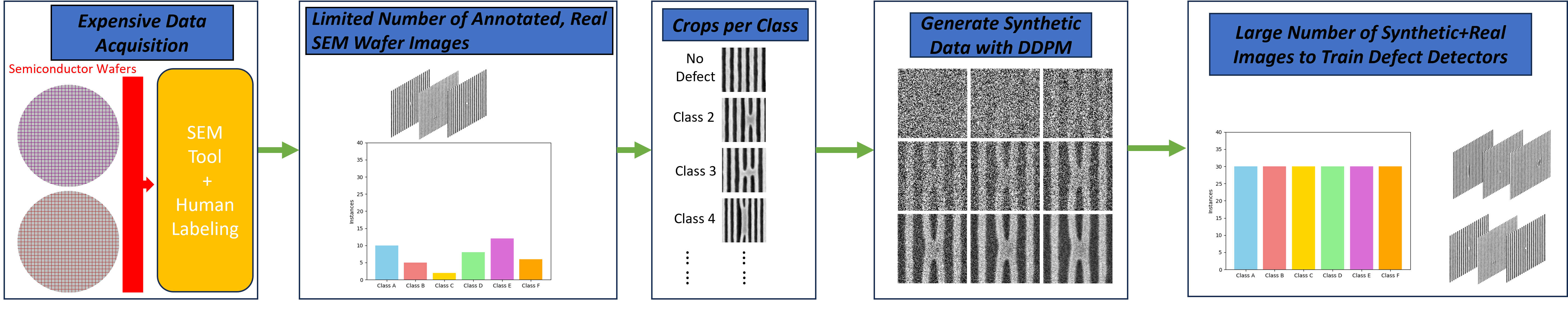}
    \caption{High level overview of the proposed approach towards synthetic SEM image dataset generation containing multi defect types.}
    \label{overview}
\end{figure*}
In earlier years, main semiconductor defect scenarios were due to particle contamination, or impurities. These defect patterns differed significantly from regular patterns, making rule-based defect inspection techniques effective for detecting such large deviations. Additionally, at those order of magnitudes, precise alignment between SEM images and non-defective reference image could be achieved, leading to reference image based defect inspection techniques achieving sufficient performance. The integration of Extreme Ultraviolet Lithography (EUVL) into the semiconductor manufacturing process leads to smaller pattern dimensions ($\leq$7nm). However, the most significant challenge associated with it is controlling stochastic defects in this nano-scale range, as most defect inspection tools lack sensitivity at this level \cite{de2014impact}. These defects are caused by unavoidable physical phenomena, such as photon shot noise, which becomes more relevant as the pattern shrinks. Thus, an extensive investigation into the relationship between process parameters and stochastic defectivity is required for every IC node during its development. This investigation is, in part, enabled by accurate defect detection and classification. However, due to i) decreased pixel-sized defect patterns, ii) increased SEM imaging noise, and iii) increased difficulty in reference image alignment, the stochastic defects at these advanced nodes cannot be detected by the previously mentioned rule-based methods \cite{rulesfail}. In contrast, ML-based techniques have achieved adequate detection rates for inspecting semiconductor wafer SEM images, even at advanced EUVL nodes \cite{dey2022deep}.

Cheon et al. \cite{cheon2019convolutional} first demonstrated superiority of Convolutional Neural Networks (CNN) approach on SEM-based semiconductor defect inspection compared to rule-based or other ML-based approaches. While this work investigated detection of particle contamination defects (surface level), performance of CNN-based approaches has also been demonstrated for nano-scale, stochastic defect inspection at the EUVL node by Dey et al.\cite{dey2022deep}. 
Moreover, ML approaches are also to improve wafer processing times, as Lorusso et al. \cite{kondo2021massive} demonstrated ML-based super resolution supports significant SEM scanning time reductions in defect inspection SEM, since it allows lower resolution SEM settings.

Ahn et al. \cite{augm} confirmed that training on both simulated and real SEM images, with augmentation strategies, results in the best defect inspection performance. This demonstrates that models benefit from a higher number of training images compared to the available real SEM images. They also note that current simulation methods do not accurately model the line-edge roughness and noise parameters of more advanced nodes. As IC pattern dimensions shrink further, several challenges such as edge \cite{edge}, or surface \cite{surface} effects persist in accurate SEM wafer imaging models. Additionally, the introduction of GPU-based computational resources has allowed significant improvements in simulation times \cite{nebula}. However, all SEM simulation approaches still assume each electron follows an independent trajectory. Introducing multi-body modeling for these electrons necessarily results in exponential increases in required computation time, even in the case of GPU-based computation. Due to these challenges, accurate SEM modeling is yet to be achieved, as discussed in Arat et al. \cite{sensitivity}.

Additionally, some defect types appear less frequently in training data, leading to class imbalance problems. While De Ridder et al. \cite{de2023semi} proposed a weighted sampling strategy with data augmentation to reduce the impact of class imbalance, the strategy was unsuccessful. Thus, data augmentation alone seems insufficient for solving the data availability problem in semiconductor defect inspection, illustrating the need for a method to generate novel training data without requiring additional SEM wafer scans.

To address this issue, Wang et al. \cite{wang2021defect} employed a style transfer Generative Adversarial Network (GAN) to transform defect-free wafer images into defective ones. While their approach notably enhanced defect classification performance, it was unable to generate defect-less patterns. Consequently, it could not produce large images suitable for training defect detection, as their method focused solely on classification improvement. López de la Rosa et al. \cite{LOPEZDELAROSA2022109743} proposed fine-tuning ResNet50 using a grid search algorithm for hyperparameter optimization, and also applied data augmentation techniques to the originally imbalanced semiconductor SEM dataset. With these training methods, a 3.74\% improvement in F1 score was achieved on the classification task, compared to ResNet 50 trained without hyperparameter tuning and data augmentations.

In recent years, significant progress has been made on ML-based image generation\cite{gui2021review, ho2020denoising, song2020denoising,fan2023reinforcement}. Furthermore, these developments have successfully addressed low data availability issues in the medical domain \cite{khader2023denoising}.
Specifically, Denoising Diffusion Probabilistic Models (DDPMs) \cite{ho2020denoising} have caused significant improvement in image generation-related tasks \cite{croitoru2023diffusion}. During training, DDPMs learn to reverse a diffusion process that incrementally adds Gaussian noise to sample images. As the final result of this diffusion process is pure noise, a trained DDPM can transform pure Gaussian noise into images similar to the data on which it was trained.

Hence, in this research work, we formulate a DDPM-based framework to generate full-scale SEM wafer images at advanced IC pattern dimensions. Our proposed approach generates synthetic SEM images with intended stochastic defect classes, including multiple instances of these defect classes, closely resembling real resist-wafer SEM images collected from SEM imaging tools. This is achieved while preserving the characteristics such as Line Edge Roughness (LER) and Line Width Roughness (LWR) of real images, with the goal of enhancing defect detection performance.

\section{Methodology}

\subsection{Proposed Diffusion-based Approach}

Due to their success in numerous other applications, and flexible usage, we investigate the potential of DDPM as a generative tool to solve the problem of low data availability in semiconductor defect inspection application. Our proposed approach does not train the diffusion model on the real SEM wafer images directly. Instead, various small patches are extracted from the original image. Each patch has a class label as either the defect type present inside the patch, or background (defect-free). The DDPM model is then trained in a class conditional manner on these patches. Fig.\ref{overview} depicts our proposed framework towards generating synthetic realistic SEM image dataset containing multiple defect types.

\begin{figure}[t!]
    \centering
    \includegraphics[width=0.6\linewidth]{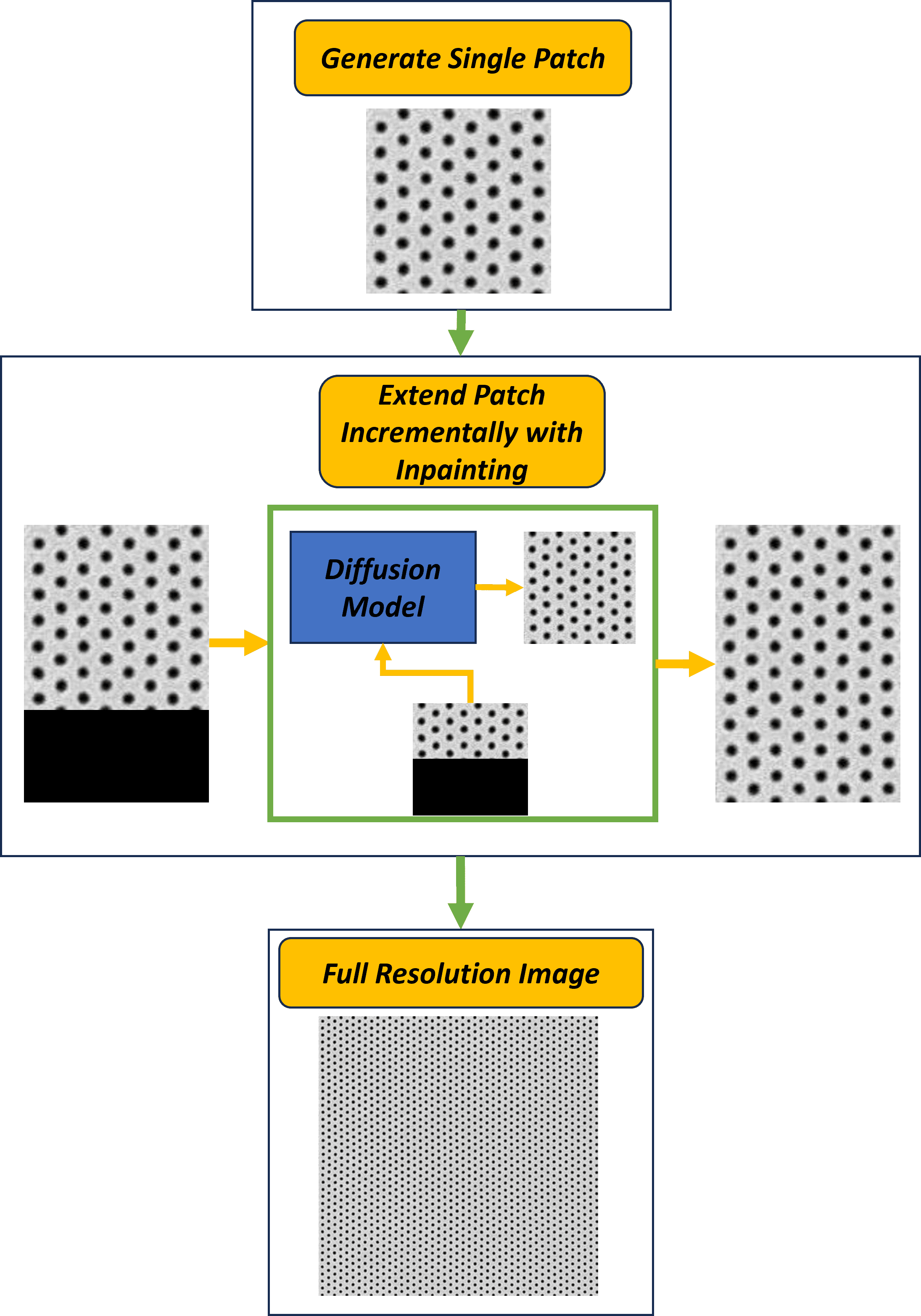}
    \caption{Proposed approach to generate full-size, defect-free SEM image (archetype) using patch based method}
    \label{method}
\end{figure}

After training, synthetic images are generated using an inpainting procedure. First, the method displayed in Fig.\ref{method} is used to generate full-size, defect-free, synthetic images. Afterwards, crops of these full-size images are inpainted to simulate intended defect types, resulting in the final synthetic images containing defects.

This patch-based approach offers three advantages over training directly on full-size SEM images: i) significantly reduced training time. ii) control over the number of defects in the generated images, enabling the generation of full-scale images with defect numbers not present in the real SEM dataset. iii) Training on patches results in larger datasets, thereby enhancing the learning process.

\subsection{Datasets}\label{objdat}

In this research work, we validate our proposed approach on three semiconductor SEM datasets: Hexagonal Contact-Hole arrays (HEXCH-DSA), Line-Space After Develop Inspection (LS-ADI), and Line-Space After Etch Inspection (LS-AEI). Each of these contains only real SEM wafer images, and no defects are synthetic or intentionally placed. The different defect types are demonstrated in Fig.\ref{exampledataset} for each dataset as: i) partially closed hole, missing hole, and closed patch for HEXCH-DSA, ii) gap, probable gap, bridge, microbridge, linecollapse for LS-ADI, and finally iii) thin bridge, single bridge, line collapse, multi brige horizontal, and multi bridge non-horizontal for LS-AEI.

\subsection{Diffusion Model: Implementation and Training}
The diffusion model used in this research work is as implemented by Nichol et al. \cite{nichol2021improved}, with cosine noise schedule and 1000 sample steps. On each real SEM dataset, the model is trained using a learning rate of 0.0001 until convergence. Code and model have been used from \cite{nichol2021improved}. We have added the inpainting functionality to the existing code, with implementation inspired by \cite{repaint}, although without their suggested resampling steps. We have trained the models on a default image size of 128 pixels. However, some defect types such as linecollapse or closed patch exceed this limit. Thus, we have separately trained a model instance on larger image sizes to generate these defect types. The hyperparameters of the models used can be found in Table \ref{settings}. 
\subsection{Training: Object Detection}
In section \ref{det}, defect detector is trained on three datasets (LS-ADI, HEXCH-DSA, LS-AEI) under different configurations (real, synthetic, combined). Due to its fast training time, YOLOv5n has been selected as defect detector\cite{ouryolo5} to validate the use of generated synthetic images in training object detectors. Each model is initialised from COCO pretrained weights, trained for 200 epochs with batch size of 32, and with early stopping criteria enabled. The weights with best performance on validation dataset are selected for use in hereafter mentioned experimentations. Code and all other hyperparameters of YOLOv5n are used as implemented by Ultralytics\cite{jocher2023yolo}.

\begin{table}
    \centering
        \caption{Diffusion Model Hyperparameters}
    \begin{tabular}{c c c}
    \hline
         \textit{\textbf{Image Size}}&\textit{\textbf{Number of Channels}} & \textit{\textbf{Residual Blocks}} \\ & \textit{\textbf{Channel Multiplier}} &\\
         \hline
         128&64 &3 \\
         & (1,1,2,2,4 ) &\\ \hdashline
         256 & 64 & 3 \\
         &(1,1,2,2,4,4)&\\
         \hdashline
         512 & 128 & 2 \\
         &(1,1,2,2,4,4)& \\
         \hline
    \end{tabular}
    \label{settings}
\end{table}

\subsection{Labeling of Synthetic Images}
To utilize synthetic images in training supervised defect detectors, such as YOLO, they must first be annotated/labelled. We propose labeling synthetic images by applying a defect detector already trained on real data. However, training defect detectors on synthetic data poses an additional challenge, as it may yield worse predictions compared to human annotation. This challenge arises due to two reasons: i) Synthetic images lack sufficient resemblance to the original data, and ii) Labeling errors in synthetic training data result in suboptimal learning signals, affecting performance on real test data. Not only have we tackled above mentioned challenges with our proposed approach, but we have also demonstrated in the next section how generated synthetic images and associated labeling quality improved or performed as per on defect detection task.
\section{Results}

\subsection{Qualitative Evaluation of Synthetic Images}
\begin{figure}
    \centering
    \begin{minipage}{0.9\linewidth}
    \begin{minipage}[b]{0.320\linewidth}
        \includegraphics[width=\linewidth]{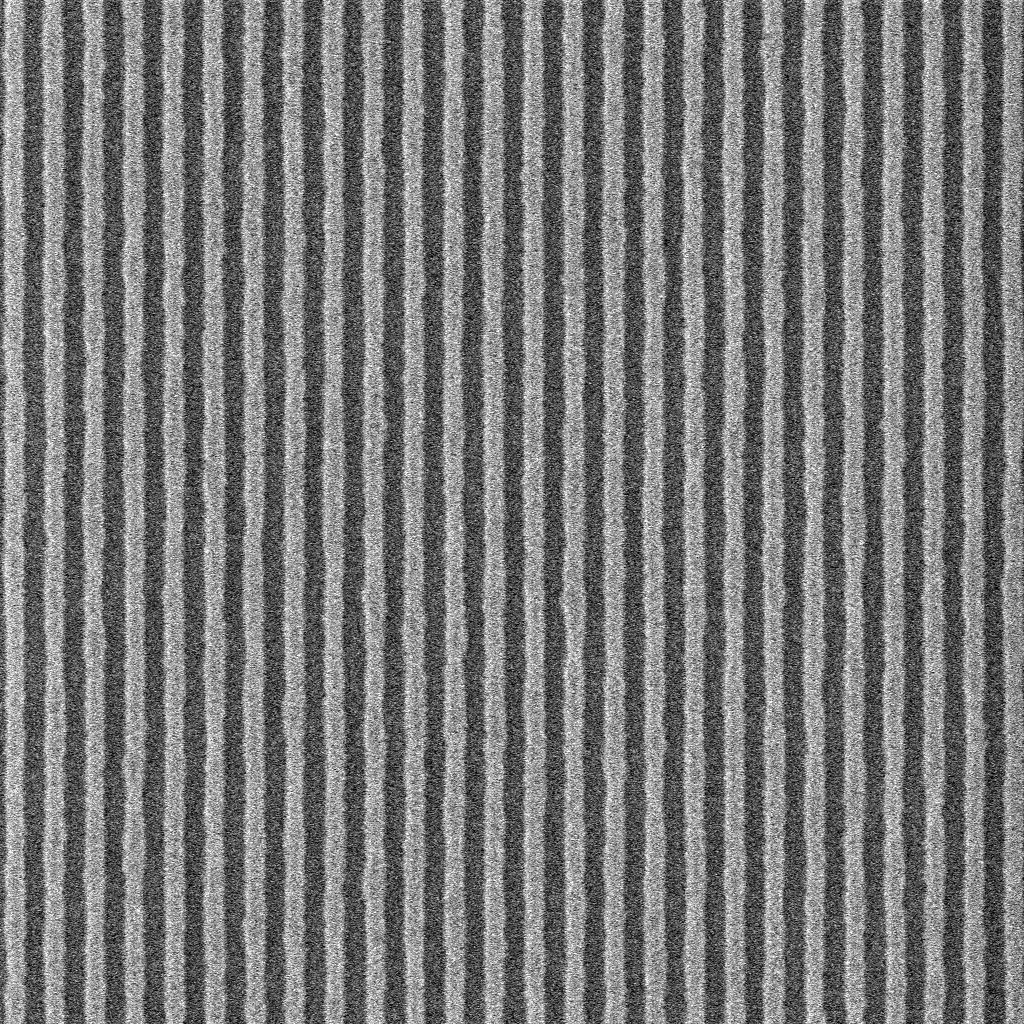}
        \caption*{(a) Real}
        \end{minipage}
        \hfill
        \begin{minipage}[b]{0.320\linewidth}
            \includegraphics[width=\linewidth]{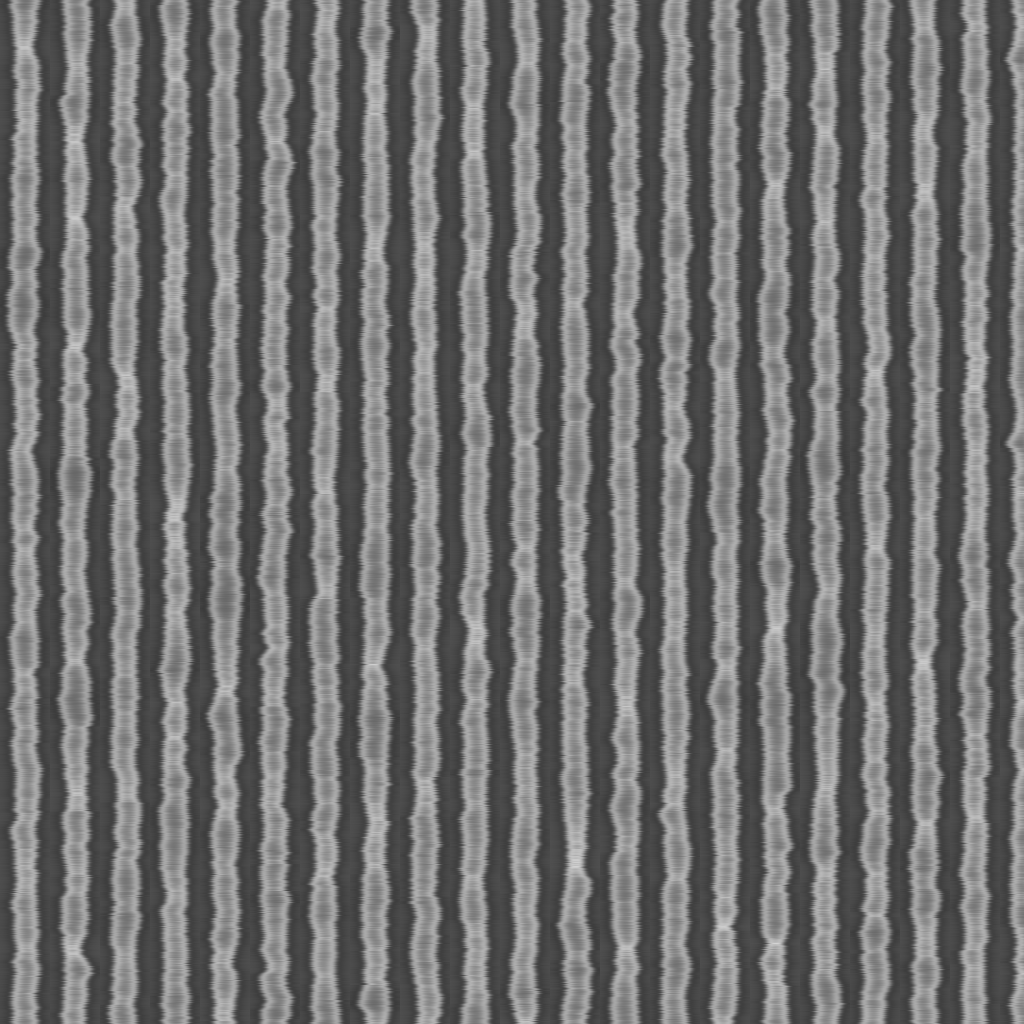}
                        \caption*{(b) Simulation}
        \end{minipage}
        \hfill
        \begin{minipage}[b]{0.320\linewidth}
            \includegraphics[width=\linewidth]{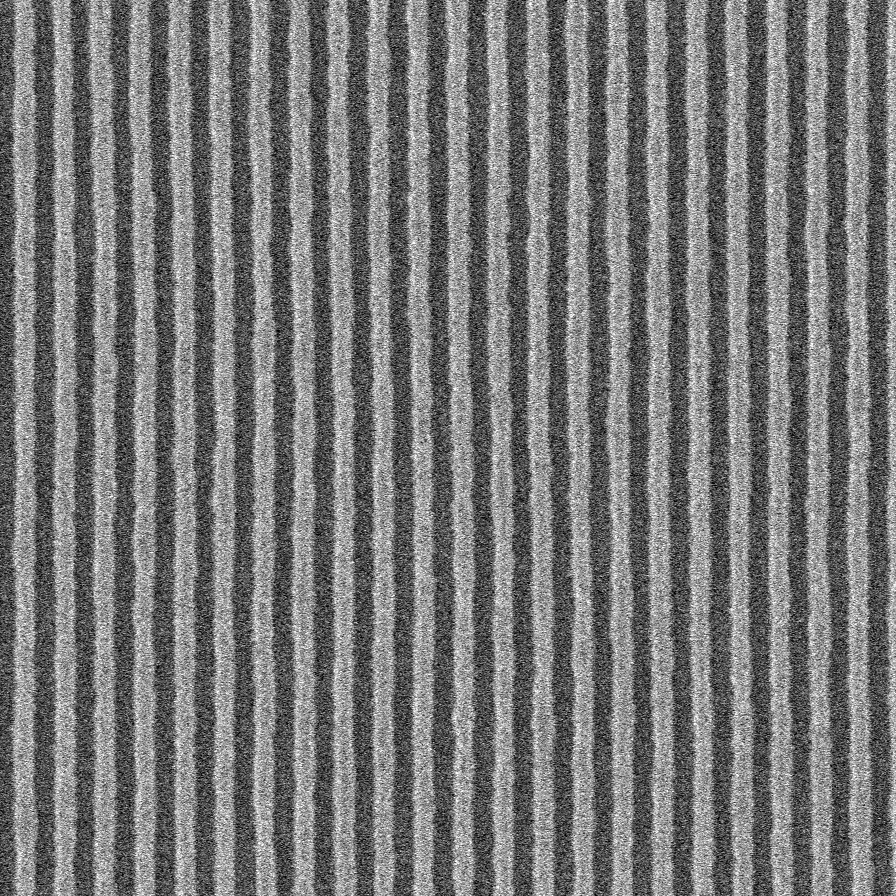}
            \caption*{(c) Proposed}
        \end{minipage}
        \caption{Comparison between (a) real SEM image, (b) image generated with software simulation, and (c) image generated with our proposed method. }
        \label{simdataset}
    \end{minipage}
\end{figure}

\begin{figure}[]
    \centering
    \includegraphics[width=0.75\linewidth]{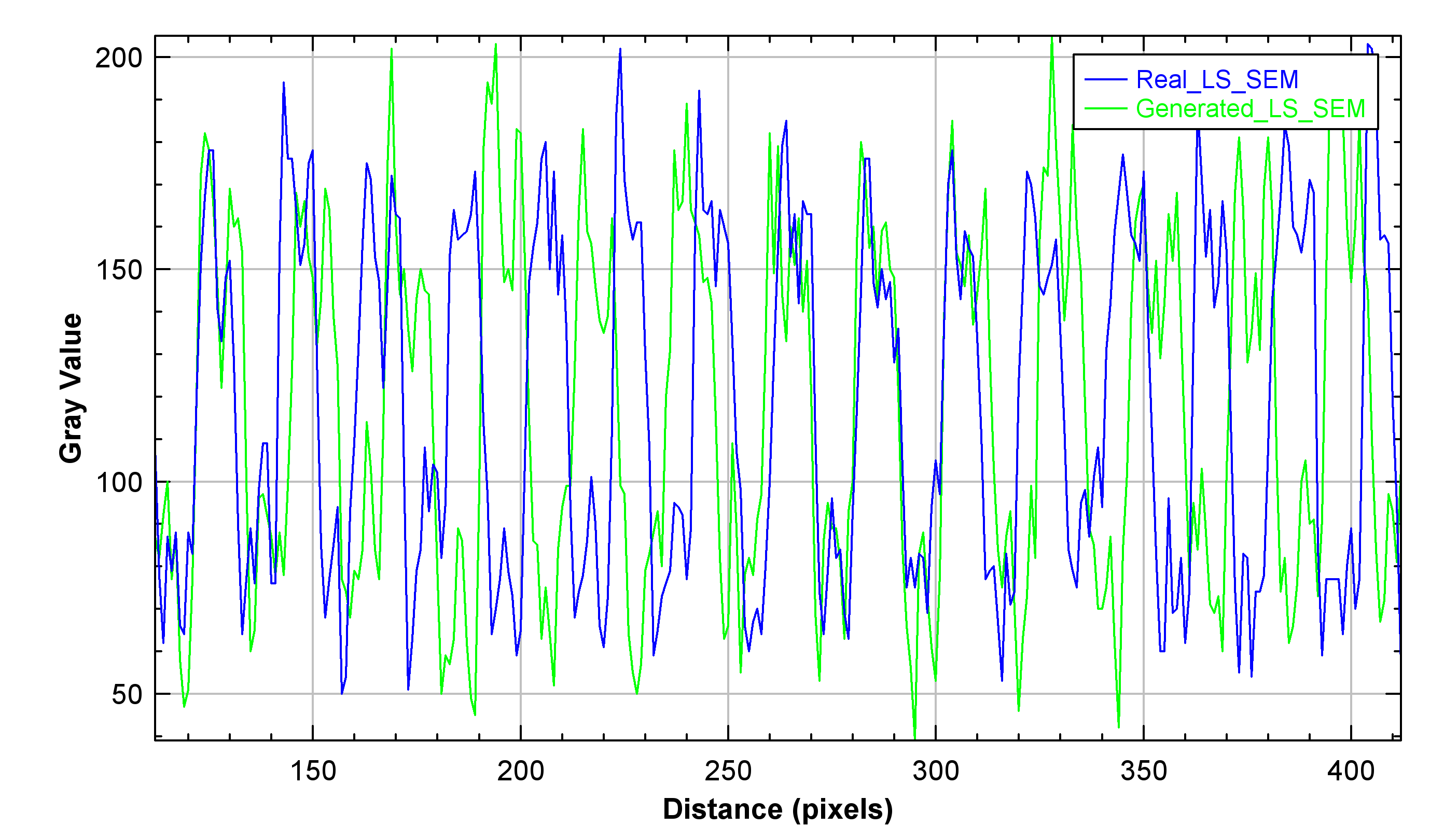}
    \caption{Linescan plot for generated (using our proposed approach) and real SEM image}
    \label{ler_comp}
\end{figure}

\begin{figure}
    \centering
    \includegraphics[width=0.6\linewidth]{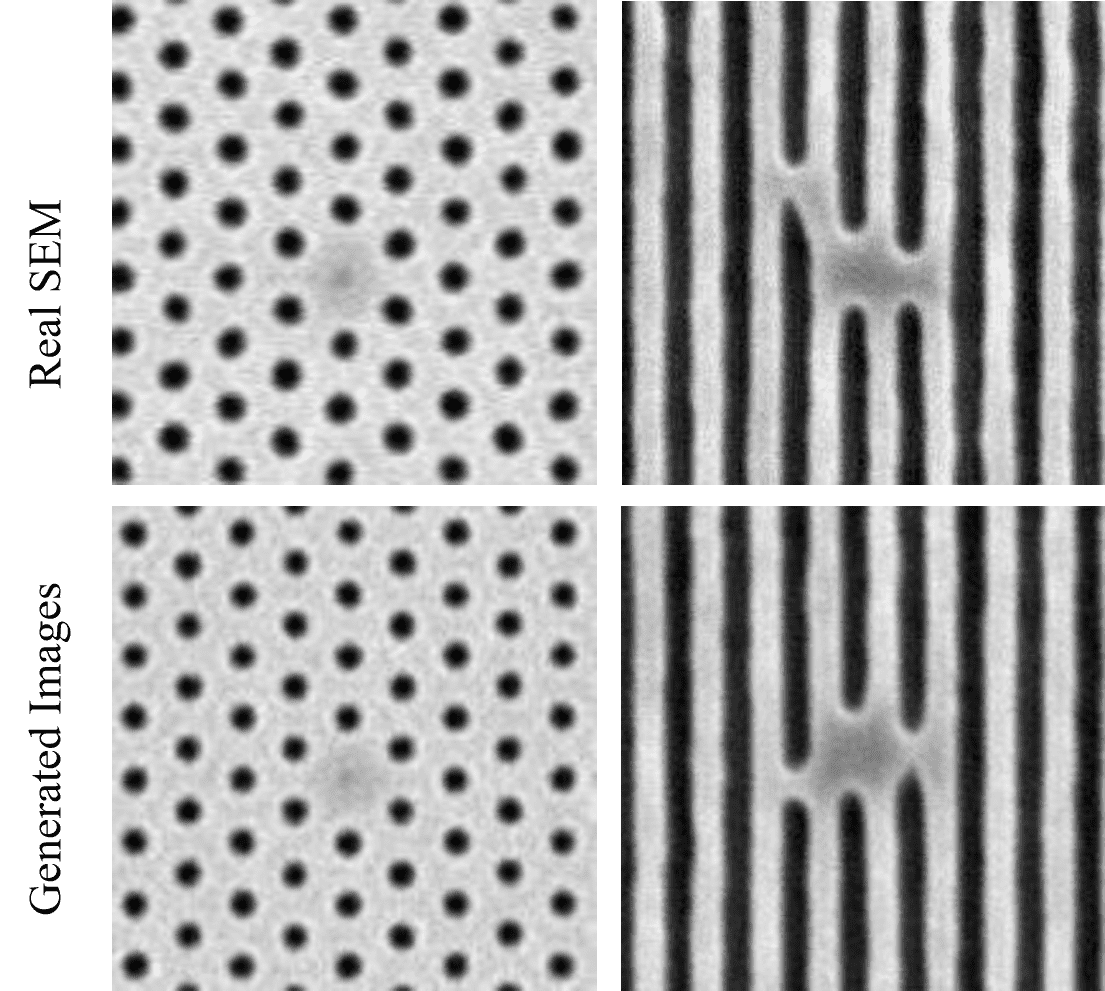}
    \caption{Examples of generated defect instances against real SEM image of same defect type}
    \label{examples_generated}
\end{figure}

\begin{figure}
    \centering
    \begin{minipage}{0.9\linewidth}
    \begin{minipage}[b]{0.320\linewidth}
        \includegraphics[width=\linewidth]{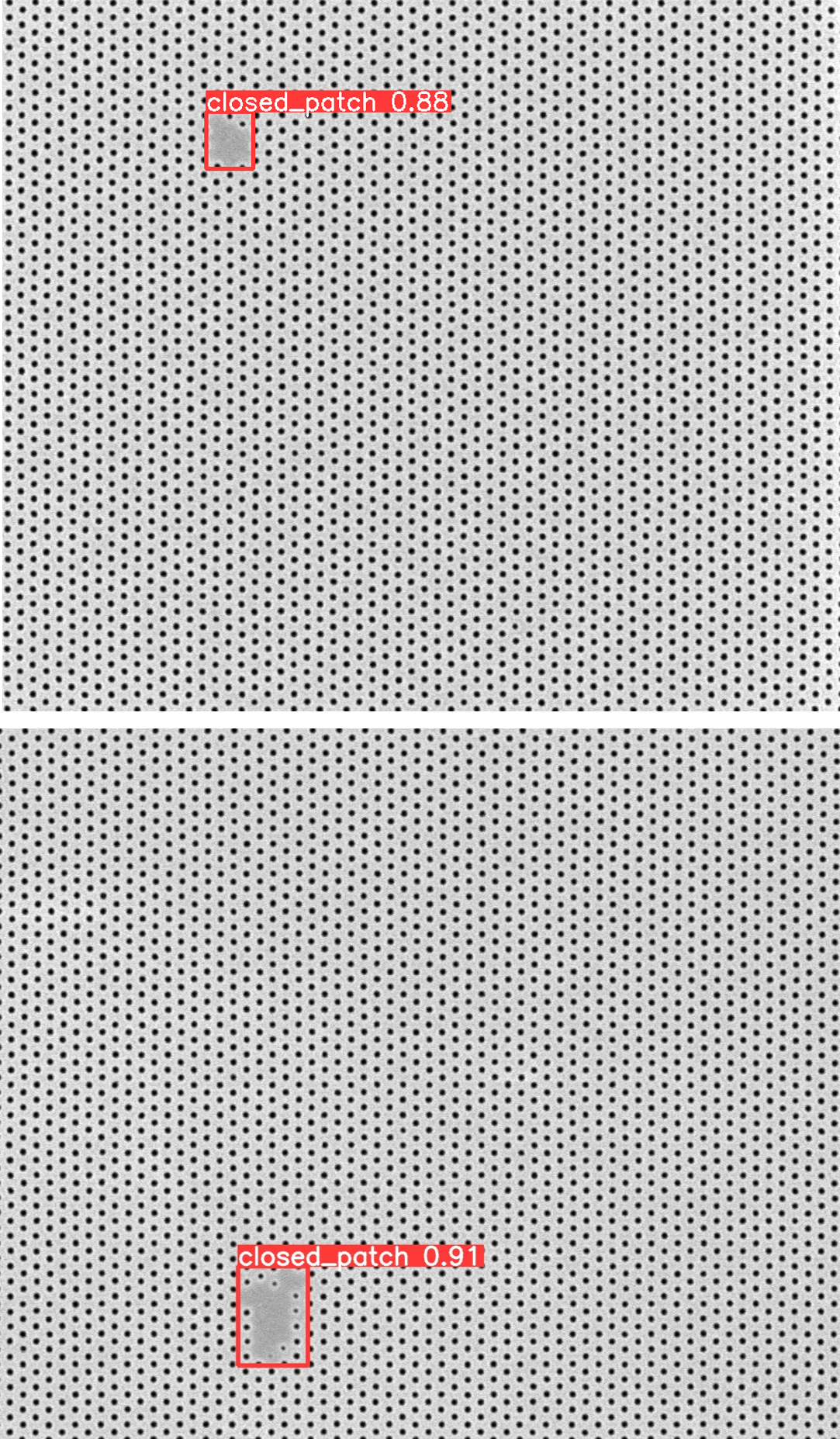}
        \caption*{(a) Closed Patch}
        \end{minipage}
        \hfill
        \begin{minipage}[b]{0.320\linewidth}
            \includegraphics[width=\linewidth]{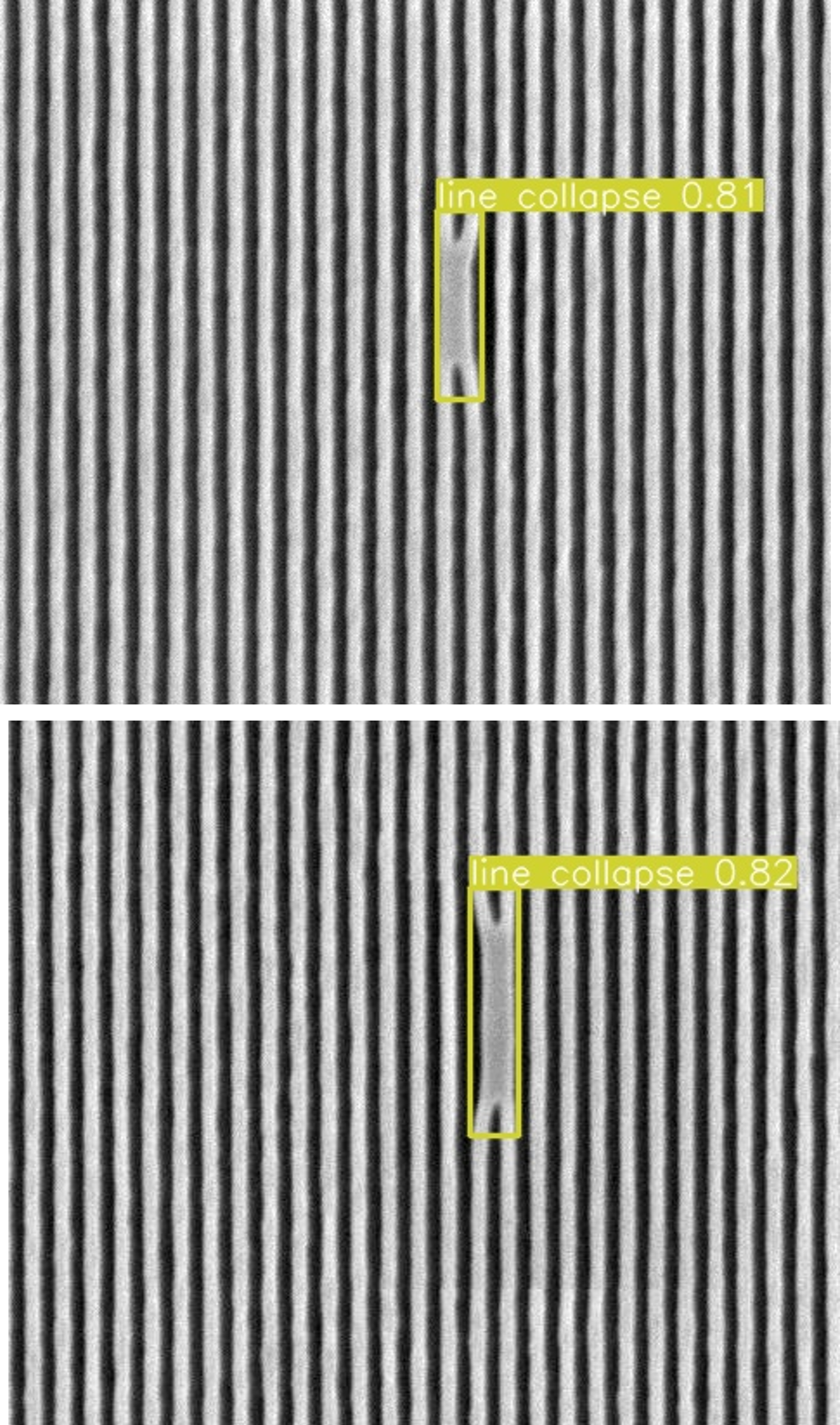}
                        \caption*{(b) Linecollapse}
        \end{minipage}
        \hfill
        \begin{minipage}[b]{0.320\linewidth}
            \includegraphics[width=\linewidth]{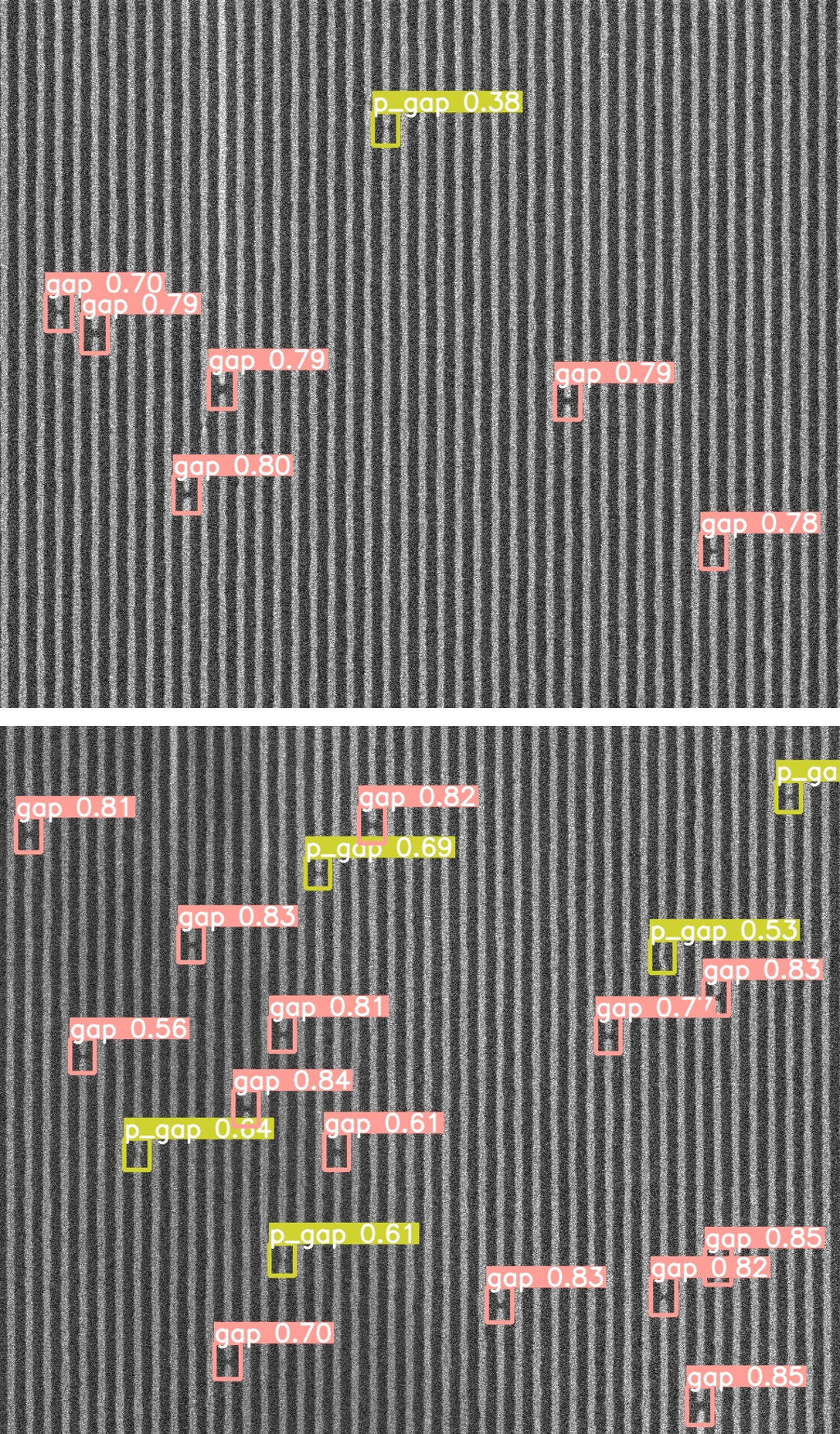}
            \caption*{(c) (p)gap}
        \end{minipage}
        \caption{Inference results for real (top), and synthetic (down) images }
        \label{inspection_results}
    \end{minipage}
\end{figure}

Synthetic images generated by the proposed DDPM-based approach are evaluated qualitatively. First, visual comparison does not yield any differentiating characteristics between synthetic defects (Fig.\ref{examples_generated}) and those obtained from SEM tools (validated with several anonymous SEM image experts). Beyond visual comparison, line-edge-roughness and critical dimension (CD) are crucial metrology parameters in semiconductor patterning, towards validating device electrical characteristics and performance. To generate synthetic or artificial SEM images, with or without defects, using conventional software such as ARTIMAGEN \cite{cizmar2008simulated, cizmar2009optimization}, it is essential to be aware of industry Best-Known-Methods (BKM) settings to comply with tool imaging conditions. Without appropriate values for parameters like pixel size, number of frames in acquisition, accelerating voltage, probe current, etc., it becomes quite challenging to generate synthetic images that closely resemble real images. Additionally, using incorrect parameter values can introduce digital artifacts into the synthetic images, rendering them unsuitable for ML model training and, consequently, compromising the preservation of original device characteristics. Contrary to this, our proposed approach generates synthetic images that closely resemble real images and preserve actual characteristics without requiring prior knowledge of BKM settings, as shown in Fig.\ref{simdataset}.
Fig.\ref{ler_comp} shows the line-scan plots of generated (by our proposed approach) and real SEM image (for Line-Space feature), which can be used to compare CD and roughness parameters between these two. Lastly, Fig.\ref{inspection_results} presents inference results on generated synthetic and real images (test set) from a defect detector trained solely on real data. Both the inference confidence and classification accuracy on semantic contexts (such as probable gap and gap), as well as the line-scan plots of original and synthetic data (for line-space feature), appear nearly identical. This strongly indicates that generated synthetic images are sufficiently similar to the original data, supporting their use in expanding the size of the semiconductor defect inspection dataset. 

\subsection{Training Defect Detectors with Synthetic Images} \label{det}

\begin{table}[]
    \centering
        \caption{Statistics of real and synthetic training datasets}
    \begin{tabular}{c| c c| c c c}
    \hline
    & & &\textbf{\textit{Real}} & \textbf{\textit{Synthetic}} & \textbf{\textit{Real+ Synthetic}}\\
    \hline
         \multirow{6}{*}{\rotatebox[origin=c]{90}{\textbf{LS-ADI}}} &\multirow{5}{*}{\textbf{Instances}}&\textit{pgap}& 315&1375 & 1690\\
         &&\textit{microbridge} & 380&1477 & 1857\\
         &&\textit{linecollapse} &550 &501 &1051\\
         &&\textit{bridge} & 238& 406& 644\\
         &&\textit{gap} & 1046& 2155&3201\\
         \cdashline{2-6}
         &\multicolumn{2}{c|}{\textbf{Total Images}}&1053& 1199&2252  \\
         \hline
         \multirow{4}{*}{\rotatebox[origin=c]{90}{\textbf{HEXCH}}} &\multirow{3}{*}{\textbf{Instances}} &\textit{cp} & 74& 210& 284\\
         &&\textit{pch} & 94 & 434& 528\\
         &&\textit{mh} & 30&591 & 621\\
            \cdashline{2-6}
         &\multicolumn{2}{c|}{\textbf{Total Images}}&174 & 420&594 \\
         \hline
         \multirow{6}{*}{\rotatebox[origin=c]{90}{\textbf{LS-AEI}}} &\multirow{5}{*}{\textbf{Instances}}&\textit{multi bridge nh} & 160& 120&280 \\
         &&\textit{multi bridge h} & 80& 143&223  \\
         &&\textit{linecollapse} &202& 248&450  \\
         &&\textit{single bridge} & 240& 245&485 \\
         &&\textit{thin bridge} &241 &354 &595 \\
         \cdashline{2-6}
         &\multicolumn{2}{c|}{\textbf{Total Images}}& 920& 932&1864 \\
         \hline
    \end{tabular}
    \label{stats}
\end{table}

Second, we demonstrate that synthetic images generated with the proposed approach can be succesfully used in training defect detectors. We have trained defect detectors with generated synthetic counterparts of each investigated dataset in two configurations as: (1) with synthetic dataset only and, (2) combined with real dataset. Statistics of real and synthetic datasets are shown in table \ref{stats}. 

\begin{figure}[]
    \centering
    \includegraphics[width=\linewidth]{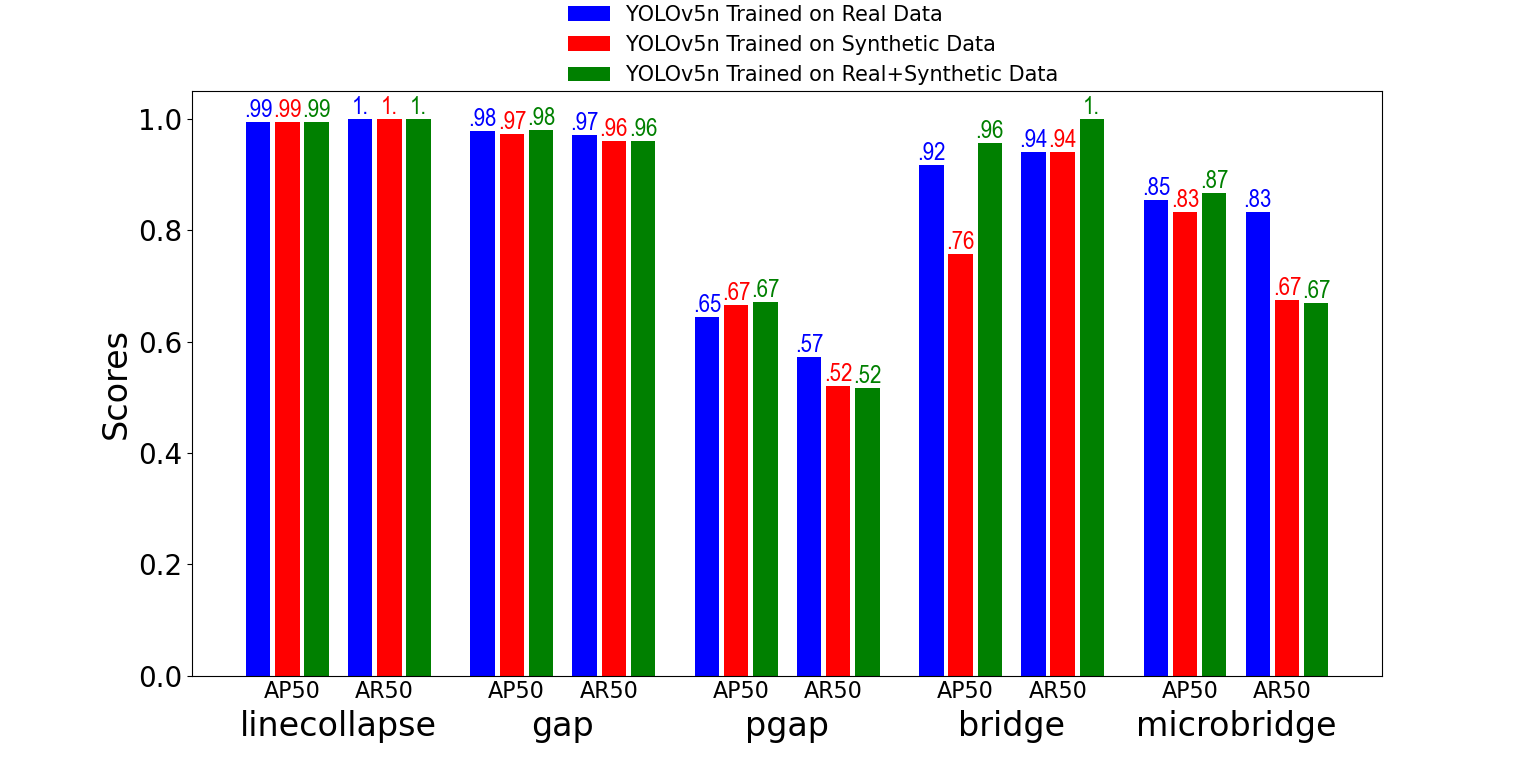}
    \caption{AP and AR scores achieved on real ADI test dataset}
    \label{adiplot}
\end{figure}

Fig.\ref{adiplot} shows the AP and AR scores per defect class on LS-ADI real test dataset, achieved by YOLOv5n model trained on either real, synthetic, or combined datasets. While some deviations and deficits are present, no major performance drops are observed when training on synthetic data. Thus, the proposed DDPM approach generates images based on the LS-ADI dataset, which can be properly utilized for training defect detectors, as no major performance deficits are encountered when switching from training on real data, to training only on synthetic data. Training only on the combination of the real and synthetic dataset does not yield any performance gains, despite the larger size of the combined dataset.

\begin{figure}[]
    \centering
    \includegraphics[width=\linewidth]{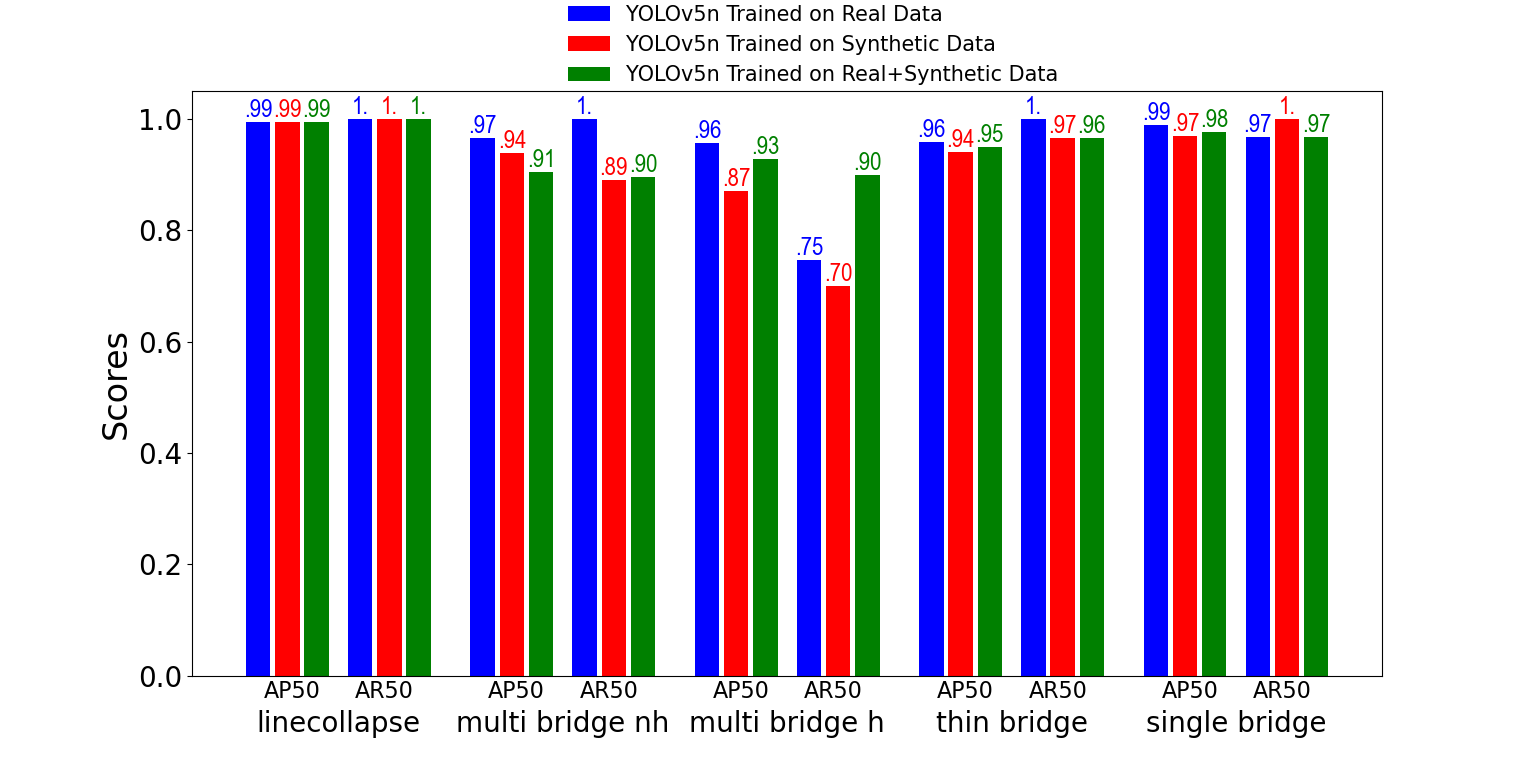}
    \caption{AP and AR scores achieved on real AEI test dataset}
    \label{aeiplot}
\end{figure}

Fig.\ref{aeiplot} shows the AP and AR scores per defect class on LS-AEI real test dataset by model trained on real, synthetic or combined datasets. A scenario similar to the LS-ADI is observed, where training on the synthetic dataset does not lead to major performance deficits.

\begin{figure}[]
    \centering
    \includegraphics[width=\linewidth]{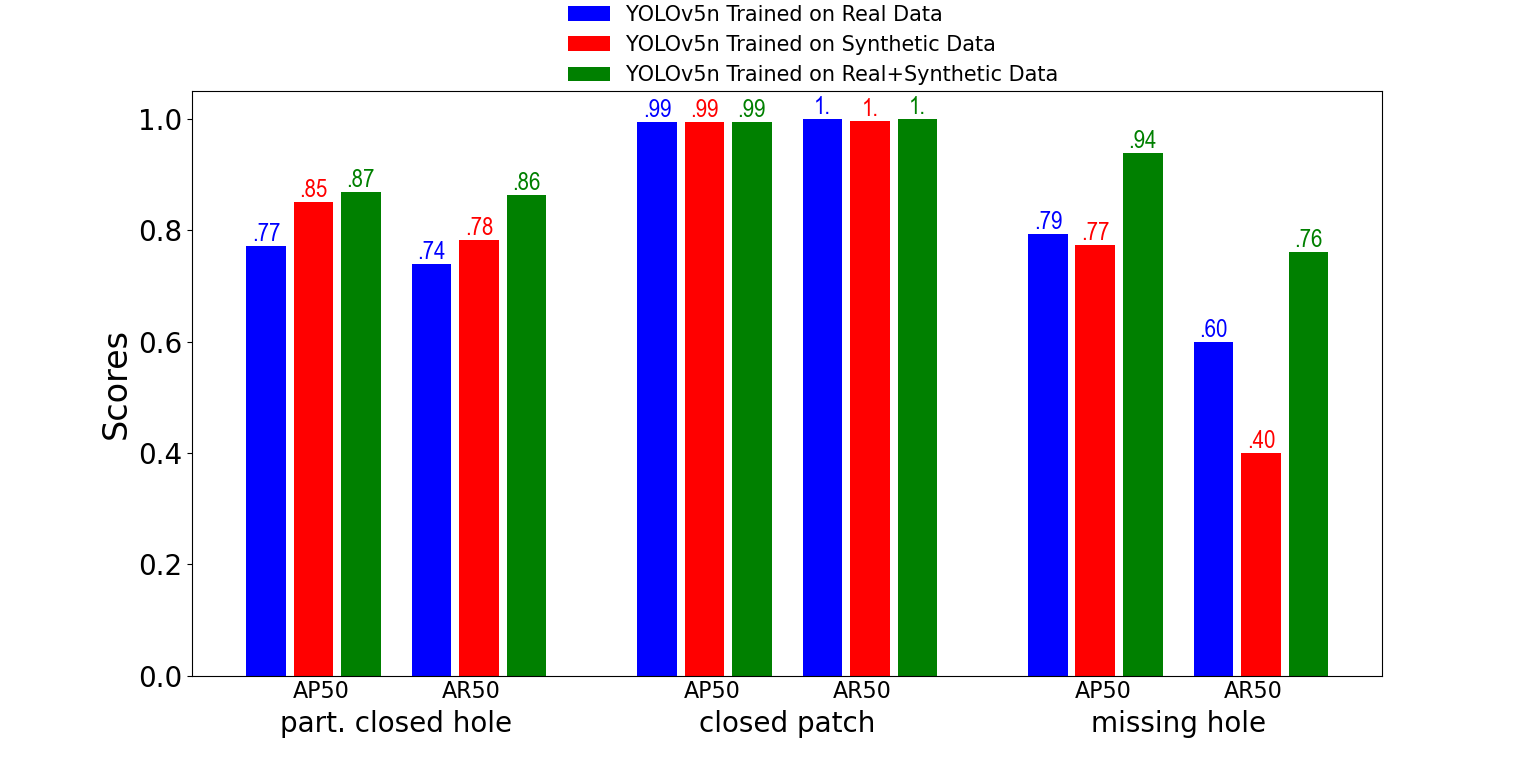}
    \caption{AP and AR scores achieved on real HEXCH-DSA test dataset}
    \label{dsaplot}
\end{figure}

Finally, Fig.\ref{dsaplot} shows the AP and AR scores on HEXCH-DSA real test dataset by model trained on the different dataset configurations. On HEXCH test data, a definite performance improvement has been observed when model is trained on the combined synthetic+real dataset. HEXCH dataset is significantly smaller in size compared to the other two datasets (table.\ref{stats}). This may explain that, while model did not benefit from training on synthetic+real data for LS datasets, the model significantly benefited for HEXCH-DSA dataset, as combining both synthetic and real data increased the dataset size (without altering real characteristics of the image/defect features) to properly learn the required defect features.  

Table \ref{performance} shows the mAP and mAR scores achieved on the different real SEM test datasets, for different training dataset configurations. This table summarises that, while training only on synthetic data does not provide clear benefits in all scenarios, it never causes significant performance drops, against training only on real dataset. This validates that the synthetic images generated by proposed method can have valid usage in training defect detection models, by replacing or combining with real SEM images datasets for different process steps.

\begin{table}[]
    \centering
        \caption{mAP and mAR scores achieved on real test data}
    \begin{tabular}{c | c| c c c}
    \hline
    \multirow{2}{*}{\textit{\textbf{Dataset (test split)}}}& \multirow{2}{*}{\textit{\textbf{Metric}}} & \multicolumn{3}{c}{\textit{\textbf{YOLOv5n with Training Data}}} \\
    & & \textit{Real}& \textit{Synthetic}& \textit{Real+Synthetic} \\
    \hline
         \multirow{2}{*}{\makecell{Real ADI-LS}} & mAP & \textbf{0.878} & 0.845 &0.839  \\
         & mAR & \textbf{0.864} & 0.819 &0.839\\
         \hline
         \multirow{2}{*}{Real HEXCH-DSA} & mAP & 0.853 & 0.873 &\textbf{0.933} \\
         & mAR & 0.78&0.85&\textbf{0.874}\\
         \hline
         \multirow{2}{*}{Real AEI-LS} & mAP & \textbf{0.973} & 0.943 &0.951 \\
         & mAR & 0.943&0.911&\textbf{0.946}\\
         \hline
    \end{tabular}
    \label{performance}
\end{table}

\subsection{Comparison to previous related work}
In this section, we illustrate how the proposed generative framework (along with the generated synthetic images) can be considered as a potential alternative data augmentation strategy to alleviate the challenges of data insufficiency and class imbalance in the semiconductor domain. Additionally, it may prove advantageous in addressing the issue of overfitting, thereby enhancing defect detection performance.

Two semiconductor defect inspection frameworks have been implemented based on previous works \cite{dey2022deep, dehaerne2023yolov8}. They are utilized to compare: (a) conventional training using only real SEM images with the application of the default data augmentation strategy[5][27], (b) training using only real SEM images without applying the default data augmentation strategy, and (c) training using a mixed dataset comprising both real and synthetic/generated SEM images generated by our proposed approach, again without applying the default data augmentation strategy. Each experiment is conducted across various architecture variants to enhance the robustness of the reported results. The best validation dataset mAP is reported for each variant and training method. We have exclusively utilized real SEM images for validation purposes.

On the LS-ADI dataset, Retinanet \cite{lin2017focal} is trained with all parameters as per Dey et al. \cite{dey2022deep}, except an increase in batch size from 2 to 16 to decrease required training time. The dataset distribution for real images (only) and (real + synthetic) images are the same as in Table \ref{stats}, Sect.\ref{det}. Table \ref{retinanet_table} presents the results achieved for three distinct backbones: ResNet 50, 101, and 152, respectively.

When comparing (b) training using only real SEM images without applying the default data augmentation strategy against (a) conventional training using only real SEM images with the application of the default data augmentation strategy from previous work \cite{dey2022deep}, we observe a significant reduction in mAP as the model size increases. This validation indicates that larger architecture variants are more susceptible to overfitting, emphasizing the importance of a data augmentation strategy to diversify the dataset encountered during training and address issues related to data insufficiency and class imbalance.

Finally, we observe that RetinaNet model trained using (c) the mixed dataset comprising both real and synthetic/generated SEM images generated by our proposed approach, without applying the conventional data augmentation strategy, achieved highest mAPs compared to previous two approaches (a) and (b), for all architecture variants (ResNet 50, 101 and 152).

For the HEXCH-DSA dataset, YOLOv8 variants are implemented with parameters according to Dehaerne et al. \cite{dehaerne2023yolov8}. The mixed dataset, containing both real and synthetic images, comprises 5425 images. The dataset consisting solely of real images remains consistent with Table \ref{stats}, Sec. \ref{det}. Table \ref{y8_dsa_table} presents the mAP values achieved for (c) training using the mixed dataset comprising both real and synthetic/generated SEM images generated by our proposed approach, without applying the default data augmentation strategy compared to  (a) conventional training using only real SEM images with the application of the default data augmentation strategy. We observe that our proposed generative approach-based data augmentation strategy, utilizing generated synthetic images, outperforms both the conventional data augmentation strategy and previous research work \cite{dehaerne2023yolov8} for most architecture variants, or performs as per (YOLOv8n and YOLOv8x).

\begin{table}[]
    \centering
        \caption{Comparitive analysis of mAP on real LS-ADI validation dataset with RetinaNet model \cite{dey2022deep}.}
    \begin{tabular}{c |  c c| c}
    \hline
     \multirow{3}{*}{\textit{\textbf{Backbone}}} & \multicolumn{3}{c}{\textit{\textbf{mAP achieved with training dataset}}} \\
\cline{2-4}
    & \multicolumn{2}{c|}{ \textit{Real}}  & \textit{Real+Synthetic}  \\
   &  \textit{Default} &  \multirow{2}{*}{\textit{No augmentation}}& \multirow{2}{*}{\textit{No augmentation}} \\
    &  \textit{augmentation} & &  \\
    \hline
         Resnet 50 & 0.739 & 0.816 &  \textbf{0.883}  \\
         Resnet 101 &  0.711 & 0.680 & \textbf{0.837} \\
         Resnet 152  & 0.743 & 0.590 &\textbf{0.852} \\
         \hline
    \end{tabular}
\label{retinanet_table}
\end{table}

\begin{table}
    \centering
        \caption{Comparitive analysis of mAP on real HEXCH-DSA validation dataset with YOLOv8 model \cite{dehaerne2023yolov8}.}
    \begin{tabular}{c | c|C{2.85cm}}
    \hline
     \multirow{3}{*}{\textit{\textbf{YOLOv8 Variant}}} & \multicolumn{2}{c}{\textit{\textbf{mAP achieved with training dataset}}} \\
\cline{2-3}
    &  \textit{Real}  & \textit{Real+Synthetic} \\
    &  \textit{Default augmentation}& \textit{No augmentation} \\
    \hline
         n & \textbf{0.930}  & 0.924  \\
         s &  0.916 &  \textbf{0.929} \\
         m  & 0.910 & \textbf{0.931} \\
        l & 0.888 & \textbf{0.953}\\
       x & \textbf{0.917} & 0.910\\
\hline
\textit{Average} & 0.91 & \textbf{\textit{0.93}} \\
\hline
    \end{tabular}
\label{y8_dsa_table}
\end{table}

\begin{figure}[t]
    \centering
    \includegraphics[width=\linewidth]{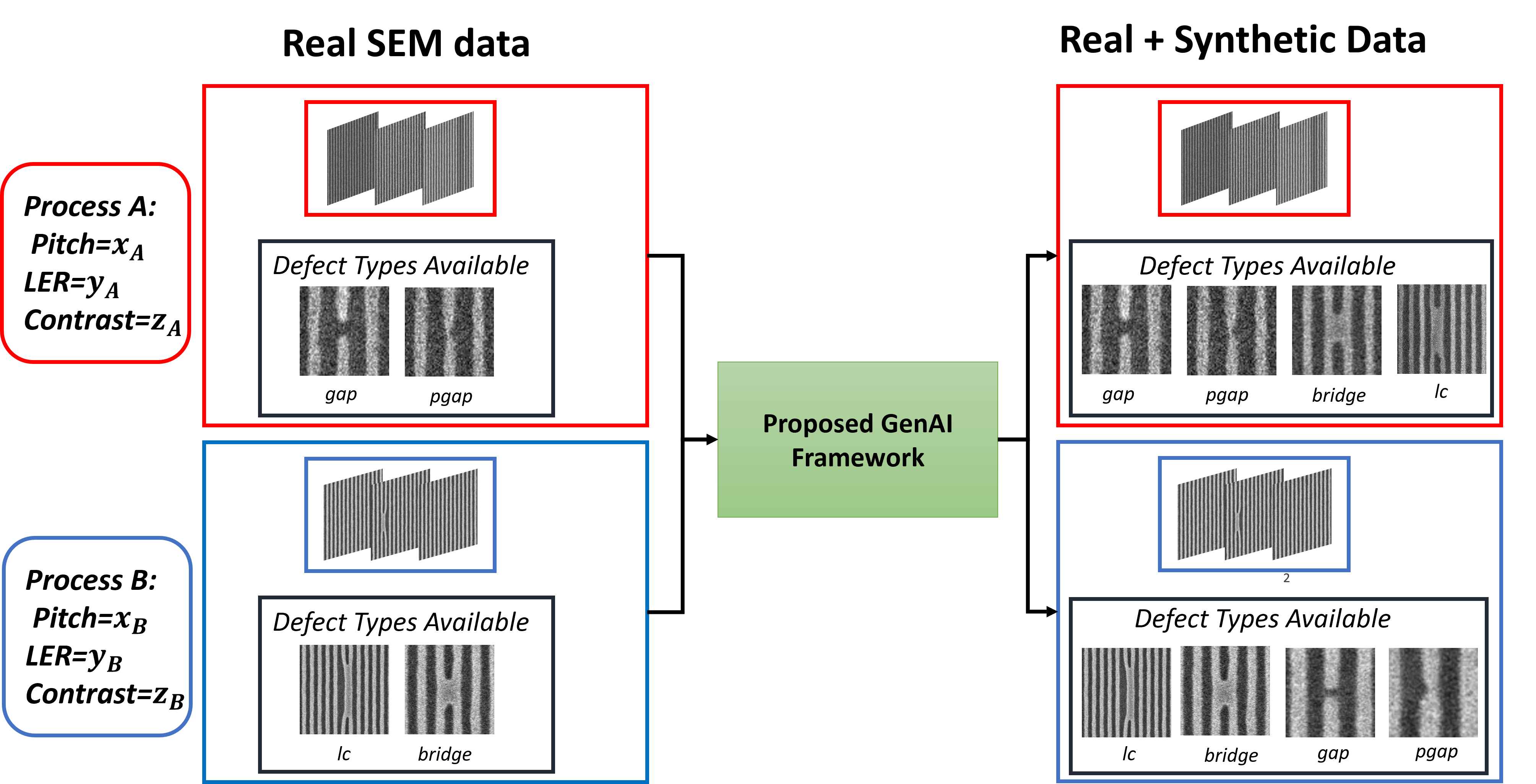}
    \caption{Proposed framework can generate defect types outside its typical process parameters, to prepare defect detectors for unexpected scenarios.}
    \label{transfer}
\end{figure}

\subsection{Defect Transfer}
SEM wafer images can vary significantly depending on factors such as dose/focus used, design geometrical patterns, critical dimension, resist profiles, or underlayers. Consequently, significant numbers of SEM images have to be acquired and labeled for each set/combination of process parameters to train a defect detection model. Furthermore, as defect types are stochastic in nature, two challenging scenarios may occur quite frequently, as (i) a given defect type/class has a very small probability to occur in that process while training defect detection models (class imbalance), or (ii) relevant defect SEM images dataset to train a model is not just rare and noisy, but also very expensive to get (limited training dataset size).

In these cases, without sufficient images of certain defect types available at a certain process step, deploying an industry-compliant ML-based defect detection framework may be problematic, as overall model convergence can not be guaranteed towards generalizability and robustness due to model's under-fitting for those defect type's features.

To mitigate this, we have examined whether the proposed approach can generate instances outside of the extent of corresponding defect type's typical process context. In this way, the proposed generative model can be trained on different processes and their associated defect types concurrently. Afterwards, defect instances can be generated for a process where the given type has not been encountered yet (or encountered in limited numbers). This generated dataset can then be used to train ML-based defect detectors towards detecting the given defect types in the new environment. Our proposed approach is demonstrated in Fig.\ref{transfer}

The proposed approach manages to successfully generate defect instances outside of the process parameters they were encountered in during training. However, as of now, no dataset which allows extensive investigation of this proposed "defect transfer" approach, is available to the authors. Therefore, quantitative experimentation and validation with this approach is left for future research directions.

\section{Conclusion}
In this research work, the application of DDPM has been investigated for semiconductor defect detection. First, a patch-based approach was formulated to generate full-size SEM wafer images. Afterwards, the quality of the generated synthetic SEM images is examined in comparison to the metrology specifications of real images. Line-scan plots of real and synthetic images were compared and showed no significant differences in relevant parameters. Subsequently, the defect detector model was trained on these generated synthetic images under different conditions to further validate the applicability of synthetic images within a robust defect inspection framework. This addresses industrial challenges, such as class imbalance and data insufficiency, especially for the stochastic defect dataset in real resist wafer images. This dataset is not only rare and noisy but also very expensive to collect, making synthetic data valuable in facilitating the training of machine learning models. We have demonstrated significant performance improvement for both the HEXCH-DSA dataset and the LS dataset in both process steps, namely ADI and AEI, while using synthetically generated images during training. Notably, training on combined real and synthetic data improved mAP by 6.9\% on HEXCH-DSA and improved AP on missing hole defect class by 19\%. Finally, it is demonstrated that the proposed approach can generate defect types outside their typical environment, and a framework for 'pattern transfer' is proposed.

\bibliographystyle{ieeetr}
\bibliography{reference}

\newpage

\end{document}